\pdfoutput=1
\PassOptionsToPackage{table,dvipsnames}{xcolor}

\documentclass[11pt]{article}

\usepackage[]{EACL2023}

\usepackage{times}
\usepackage{latexsym}

\usepackage[T1]{fontenc}

\usepackage[utf8]{inputenc}

\usepackage{microtype}

\usepackage{inconsolata}

\usepackage{custom}

\title{Do LLMs Truly Benefit from Longer Context in Automatic Post-Editing?}

\author{
Ahrii Kim$^{\dagger\ddagger}$\thanks{Corresponding author.}
\hspace{4em} Seong-heum Kim$^{\dagger}$ \\
$^{\dagger}$ Soongsil University \\
$^{\ddagger}$ AI-Bio Convergence Research Institute \\
\texttt{\href{mailto:ahriikim@ssu.ack.kr}
{\{ahriikim,seongheum\}@ssu.ac.kr}}}

\begin{document}
\maketitle

\begin{abstract}
Automatic post-editing (APE) aims to refine machine translations by correcting residual errors. Although recent large language models (LLMs) demonstrate strong translation capabilities, their effectiveness for APE—especially under document-level context—remains insufficiently understood. We present a systematic comparison of proprietary and open-weight LLMs under a naive document-level prompting setup, analyzing APE quality, contextual behavior, robustness, and efficiency.

Our results show that proprietary LLMs achieve near human-level APE quality even with simple one-shot prompting, regardless of whether document context is provided. While these models exhibit higher robustness to data poisoning attacks than open-weight counterparts, this robustness also reveals a limitation: \textbf{they largely fail to exploit document-level context for contextual error correction.} Furthermore, standard automatic metrics do not reliably reflect these qualitative improvements, highlighting the continued necessity of human evaluation. Despite their strong performance, the substantial cost and latency overheads of proprietary LLMs render them impractical for real-world APE deployment. Overall, our findings elucidate both the promise and current limitations of LLM-based document-aware APE, and point toward the need for more efficient long-context modeling approaches for translation refinement.\footnote{\faGithub{}\url{https://github.com/trotacodigos/Longform-Eval.git}}
\end{abstract}

\section{Introduction}
Large language models (LLMs) have demonstrated remarkable progress across a wide range of tasks, including reasoning, summarization, code switching, and machine translation \citep{li:24-retrieval,potter-yuan-2024-llm,laban:24-summary,kocmi:24-wmt,huang2024survey,zhang2025survey}. In particular, the quality of LLM-based translation, when evaluated at the sentence level, has been reported to approach or even surpass human performance \citep{kocmi:24-wmt,proietti-perrella-navigli-2025-human,zhang2025llm_literary}. Regardless of whether this claim holds universally, it is clear that the quality of model-based translation has reached a new peak.

Human translation from scratch, on the other hand, is a time-consuming and cognitively demanding task. Reported throughput for professional translators varies across language pairs and domains, but \citet{toral:19} estimates about 503 sentences per hour—an amount that GPT-4o can process within 2–6 \textit{seconds} (about 80–230 tokens per second). Post-editing (PE), where human translators correct residual errors in machine translation (MT), has been proposed to alleviate this inefficiency \citep{informatics7020012,zouhar-etal-2021-neural,sarti-etal-2022-divemt}. However, PE raises persistent concerns, such as the production of "translationese" or unnatural phrasing resulting in ``post-editese'' \citep{toral:19-post,li-etal-2025-lost}. More importantly, since sentence-level MT often fails to account for discourse-level coherence, PE outputs may require additional cognitive effort from humans \citep{sanchez:16-machine,ALVAREZVIDAL2023100125,wang:23:cognitive}.

With the advent of long-context LLMs (capable of processing $\approx$ 32k–108k tokens at once), it is timely to reconsider the role of PE—particularly the potential of automatic post-editing (APE) that mimics \textit{human reading behaviour} by considering full-document context. Recent studies have explored LLM-based APE and reported encouraging results \citep{raunak:23-leveraging,koneru-etal-2024-contextual,velazquez:25-langmark,deoghare:25-giving}, even in low-resource settings \citep{deoghare:24-together}. Nevertheless, comprehensive assessments of context-aware APE and the way LLMs utilize document context remain largely unexplored.

In this work, we address this gap by conducting a comparative analysis of LLM-based APE with a particular focus on document-level translation. We systematically evaluate proprietary and open-weight models under a naive full-document prompting setup, examining their editing behavior, contextual sensitivity, robustness, and efficiency.

Our main contributions are as follows:
\begin{itemize}[noitemsep]
    \item We show that LLM-based APE can \textbf{reach human parity under simple prompting}, yet provides no consistent gains over segment-level APE when document-level context is introduced, revealing \textbf{fundamental limitations in current context utilization}.
    
    \item Through controlled experiments, we demonstrate that open-weight LLMs are highly sensitive to long and noisy contexts, suffering substantial degradation under \textit{data poisoning attacks} \citep{zhao2025datapoison,shi2023datapoison}, while proprietary models exhibit greater robustness but limited contextual awareness.
    
    \item We find that most APE edits produced by LLMs are \textbf{paraphrastic in nature} and poorly captured by surface-level automatic metrics, reinforcing the necessity of human evaluation, as also emphasized by \citet{mathur-etal-2020-tangled}.
    
    \item Finally, we quantify the significant cost and latency overheads of document-level LLM-based APE, highlighting a critical gap between observed performance and practical deployability.
\end{itemize}
\section{Related Work}
Appendix~\ref{appx:works} provides a comprehensive explanation.

\subsection{Long-Context Modeling in LLMs}
Recent advances in long-context LLMs have expanded input lengths from thousands to hundreds of thousands of tokens by improving memory handling, attention mechanisms, and training strategies.
These approaches can be broadly categorized into: (1) \textit{architectural optimizations}, such as sparse or sliding-window attention and KV-cache reuse \citep{yen-etal-2024-long,zahirnia2025ett,liao2025E2LLM}; (2) \textit{retrieval- and memory-augmented models} that dynamically retrieve context beyond fixed windows \citep{he-etal-2025-hmt,chen2025ccat,vasylenko2025sparse-attention}; and (3) \textit{data-centric approaches}, including long-context fine-tuning and synthetic context expansion \citep{tian-etal-2025-untie,zhao2024longskywork}.
We build upon these advances to examine whether long-context capability benefits APE.

\subsection{LLM-Based APE}
Instruction-tuned LLMs have recently shown strong potential in APE without task-specific fine-tuning.
\citet{ki-carpuat-2024-guiding} and \citet{lu-etal-2025-mqmape} leveraged MQM-based signals to guide LLMs for error correction, while \citet{kim:25:multiagent} employed a multi-agent framework where translation and PE were performed separately under Rubric-MQM \cite{kim-2025-rubric} supervision.
LLMs have also been used to assist human post-editor in a real-world setup \citep{yuksel2025ape}.
However, most studies focus on sentence-level APE and neglect document-level context; only \citet{li-etal-2025-enhancing-large} explored pseudo-document training, without analyzing contextual behavior or efficiency—our primary focus.

\subsection{Meta-Evaluation of APE}
Research on evaluating how LLMs behave as post-editors remains limited.
\citet{raunak:23-leveraging} assessed GPT-4 on WMT22, noting both quality gains and hallucination risks.
\citet{velazquez:25-langmark} and \citet{deutsch-etal-2025-wmt24} introduced large-scale human-annotated APE datasets, enabling multilingual evaluation.
Yet, none of these studies analyze \textit{how} LLMs exploit document context.
We fill this gap by comparing proprietary and open-weight LLMs in document-level APE, analyzing their quality, efficiency, and contextual awareness.

\section{Approach}
\label{sec:approach}
\subsection{Problem Formulation}
Our goal is to enable a model to perform post-editing by leveraging document-level context, similar to how human translators work.  

Let a source document be $D_S = \{S_1, S_2, \dots, S_N\}$ consisting of $N$ segments.  
A translation model or human produces a target document  
$D_T = \{T_1, T_2, \dots, T_N\}$.  

A PE function $\mathcal{P}$ revises the translation output, yielding
\[
 \mathit{PE}_T = \mathcal{P}(D_S, D_T).
\]

The conventional APE process is performed at the sentence level (\textcolor{blue}{\APEseg}), considering only one source–translation pair at a time:
\[
 \mathit{PE}_T = \{\, \mathcal{P}(S_i, T_i) \,\}_{i=1}^N.
\]

In contrast, we consider a document-level approach (\textcolor{blue}{\APEdoc}) where additional context from $D_S$ and $D_T$ is injected into the PE process:
\[
 \mathit{PE'}_T = \{\, \mathcal{P}(S_i, T_i, D_S, D_T) \,\}_{i=1}^N.
\]

Our objective is to measure the quality gap $Q_{\Delta}$ between the two PE outputs:
\[
Q_{\Delta} = Q( \mathit{PE'}_T, R^\dagger) - Q( \mathit{PE}_T, R^\dagger),
\]
where $R^\dagger$ is the human reference translation and $Q(\cdot)$ denotes a quality evaluation metric (e.g., COMET, BLEURT, or human evaluation). This formulation allows us to investigate how incorporating document context $D$ into $\mathcal{P}$ affects the quality of translation.

\subsection{Instruction}
\label{subsec:instruction}
\begin{figure}[t]
\centering

\begin{lstlisting}
You are a professional post-editor. 
Your task is to improve draft translations.
Follow these rules:
- Use the source and draft translation as the primary references.
- Source/target documents are for context only; never copy text directly.
- The final translation must be fluent, natural, and faithful to the source.
- Do not add or invent information.
- Output ONLY the corrected {tgt_lang} sentence wrapped with <pe> and </pe>. Nothing else.

[{src_lang} Source]
{src_seg}

[{tgt_lang} Draft Translation]
{tgt_seg}

[Source Document] (context only)
{src_doc}

[Target Document] (context only)
{tgt_doc}

Return the output ONLY in this format:
<pe>{tgt_lang} corrected sentence only</pe>

Example:
Source: "I have a cat."
Draft Translation: "Tengo un perro."
Expected Output: <pe>Tengo una gata.</pe>
\end{lstlisting}
\caption{Prompt for a document-aware (\APEdoc) setting. The \APEseg{} prompt is illustrated in Figure~\ref{fig:prompt_seg} in the Appendix.}
\label{fig:doc_prompt}
\end{figure}

As LLMs are prone to hallucination, we design our naive \APEdoc{} prompts with the following objectives:
\begin{itemize}[noitemsep]
    \item Ensure adequacy to the source without introducing extraneous information,
    \item Enforce uniqueness such that the output does not overlap with any sentence in the surrounding document, and
    \item Require that the result be composed solely in the target language.
\end{itemize}

\noindent Therefore, we compose our prompt as shown in Figure~\ref{fig:doc_prompt}. 
Making the model follow instructions and produce correct outputs becomes particularly challenging when the prompt length approaches the model’s context limit. First, overly long prompts may distract the model from the main task~\citep{shi2023datapoison}. Second, irrelevant content in the document can be regarded as ``data poisoning'' that distracts the model's judgment~\citep{zhao2025datapoison}. 

To mitigate these issues, we include an in-context learning (ICL) example that significantly improves instruction adherence~\citep{brown2020language}. Furthermore, we place the document context at the end of the prompt, as our preliminary experiments indicate that positioning it at the beginning often leads the model to overlook the translation task and sometimes even to reproduce the entire document verbatim.

\begin{table*}[ht]
\centering
\resizebox{0.85\linewidth}{!}{
\begin{tabular}{llcccccc}
\toprule
 & \textbf{Model} & \textbf{Params} & \textbf{Window} & \textbf{License} & \textbf{Max Tokens} & \textbf{Max Context} & \textbf{En-Ko} \\
\midrule
\logo{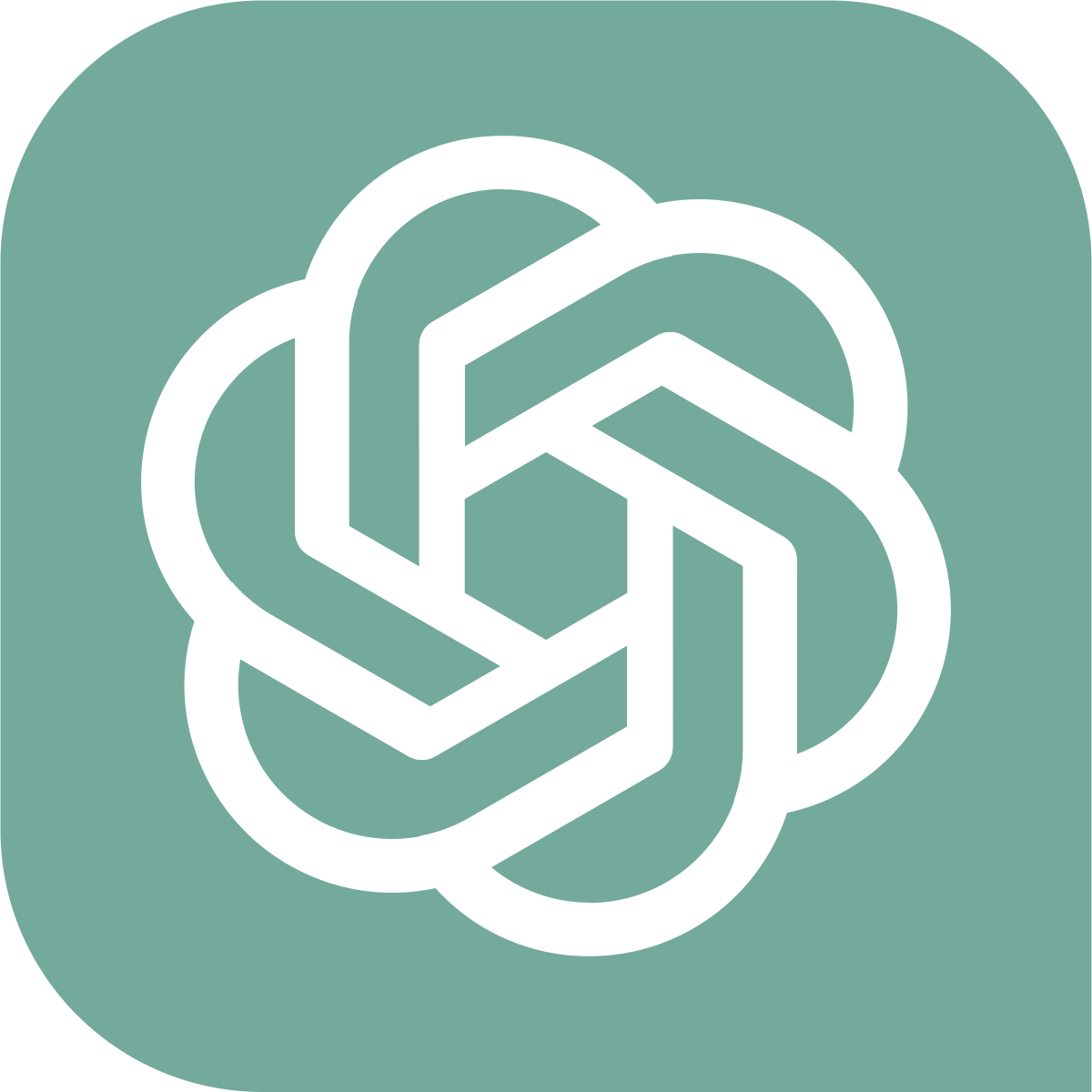} & GPT-4o & ? & 128k & Proprietary & 2,048 & -- & \textcolor{green!50!black}{\faExclamationCircle} \\
\logo{figures/gpt.png} & GPT-4o-mini & ? & 128k & Proprietary & 1,024 & -- &  \textcolor{green!50!black}{\faExclamationCircle} \\
\logo{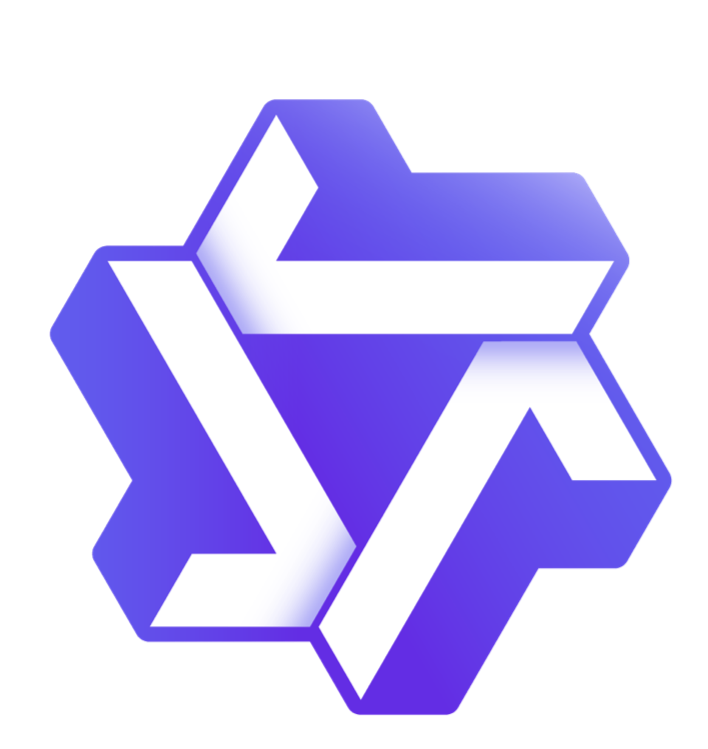} & Qwen2.5-32B & $\sim$32B & 128k & Apache 2.0 & 2,048 & 16,384 & \textcolor{blue}{\faCheckCircle} \\
\logo{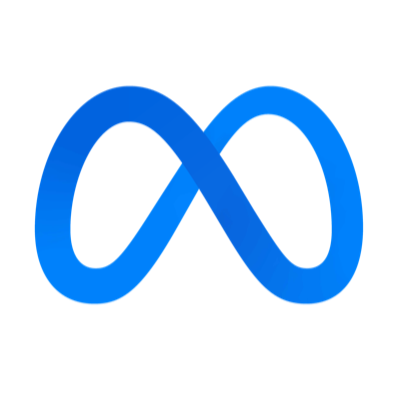} & LLaMA3-8B & $\sim$8B & 8k & Apache 2.0 & 1,024 & 8,192 & \textcolor{red!80!black}{\faTimes}  \\

\bottomrule
\end{tabular}}
\caption{Model specifications including parameter size, context window, license, and decoding parameters (\texttt{max\_token}, \texttt{max\_ctx}). The support for En-Ko language pair is also reported. 
Legend: \textcolor{blue}{\faCheckCircle}~officially supported; 
\textcolor{green!50!black}{\faExclamationCircle}~unofficial but possible; 
\textcolor{red!80!black}{\faTimes}~unsupported.}
\label{tab:model_specs}
\end{table*}

\section{Experiment}
We aim to answer the following research questions:

\begin{tcolorbox}[
  colback=gray!10,
  colframe=gray!40,
  boxrule=0.5pt,
  arc=2pt,
  left=6pt,
  right=6pt,
  top=6pt,
  bottom=6pt
]
\begin{itemize}[noitemsep]
    \item[ RQ1.] Does \APEdoc{} improve translation quality compared to \APEseg{}?
    \item[ RQ2.] If so, which aspects of translation quality benefit from the document context?
    \item[ RQ3.] How do different LLMs compare under identical experimental conditions?
    \item[ RQ4.] Is such improvement captured by automatic metrics?
\end{itemize}
\end{tcolorbox}

\subsection{Design}
Given a source document and its corresponding machine-generated translation, models are prompted to perform PE while having access to the full source and target documents as additional context. To ensure comparability across models, we adopt the same few-shot prompting template with one ICL example, placing the document context at the end of the prompt, as discussed in Section~\ref{subsec:instruction}.

\subsection{Data}
We use the WMT24++ dataset \citep{deutsch-etal-2025-wmt24}, which provides source, reference, and professionally post-edited translations. Among the available En–X language pairs, we select En–Ko, representing mid-high-resource, non-Latin language pair. Korean is an agglutinative language characterized by pro-drop phenomena, honorifics, and morphological complexity. These features make it context-dependent and suitable for assessing \APEdoc{} beyond European languages.

After removing domains with only a single document (e.g., ``canary'' and speech), the dataset comprises 7,974 segments and 59 documents, as described in Table~\ref{tab:dataset_stats} in the Appendix. Figure~\ref{fig:boxplot} summarizes the distribution of segment counts per document across domains based on the NLTK tokenizer \citep{bird2009natural}. We employ the Tukey Interquartile Range \citep{tukey:1977:EDA} (IQR) method to detect possible outliers and to illustrate data distributions. Literary texts are the longest (median 608, IQR 555–680), news documents are shorter and more homogeneous (median 153, IQR 117–207), and social documents are the shortest but most variable (median 104, IQR 60–167), with occasional outliers above 500 sentences. 
\begin{figure}[t]
\centering
\resizebox{0.9\linewidth}{!}{
\begin{tikzpicture}
\begin{axis}[
  boxplot/draw direction=y,
  xtick={1,2,3},
  xticklabels={Literary,News,Social},
  ylabel={\small Segments per document},
  width=\columnwidth,
  height=6cm,
  ymajorgrids,
]
\addplot+[
  boxplot prepared={
    lower whisker=441,
    lower quartile=554.5,
    median=608,
    upper quartile=679.5,
    upper whisker=774
  },
] coordinates {};

\addplot+[
  boxplot prepared={
    lower whisker=99,
    lower quartile=117,
    median=153,
    upper quartile=207,
    upper whisker=252
  },
] coordinates {};

\addplot+[
  boxplot prepared={
    lower whisker=10,
    lower quartile=59.5,
    median=104,
    upper quartile=166.5,
    upper whisker=325
  },
] coordinates {(3,577) (3,675)};

\end{axis}
\end{tikzpicture}
}
\caption{Distribution of segment counts per document by domain (Tukey 1.5$\times$IQR).}
\label{fig:boxplot}
\end{figure}

\subsection{Models}
We employ general-purpose LLMs with multilingual translation and PE capabilities, as well as the capacity to process long documents. Specifically, we use GPT-4o and GPT-4o-mini \cite{openai2024gpt4o} as proprietary models, and LLaMA3-8B \citep{grattafiori2024llama3herdmodels} and Qwen2.5-32B \citep{qwen25technicalreport} as open-weight models. Table~\ref{tab:model_specs} summarizes their parameter sizes, architecture types, and maximum context window lengths. We use the models in the few-shot setting with identical decoding parameters across experiments for fair comparison (\verb|temperature=0.2, top_p=0.9|). For computational efficiency, we also measure token usage and inference time overheads.

Proprietary models (GPT-4o, GPT-4o-mini) are reported to support over 50 languages across text and audio~\citep{openai2024gpt4o}, while Qwen2.5-32B explicitly covers 29+ languages including Korean~\citep{qwen25technicalreport}. As for LLaMA3-8B, Meta officially supports only eight languages, explicitly excluding Korean. In addition, its maximum context window (8k) is considerably shorter than that of the other models. We, therefore, include LLaMA3-8B as a \textit{contrastive} baseline to analyze how limited context length and unsupported language pairs affect \APEdoc{} performance. 

\subsection{Evaluation}
We assess model outputs using both automatic metrics and human evaluation, comparing them against multiple reference translations.
Specifically, we consider the following types of translations in our evaluation:\footnote{All outputs except APE are pre-supplied as part of the existing dataset}

\begin{itemize}[noitemsep]
    \item \textbf{Initial MT}: the original machine-generated translation
    \item \textbf{Human Reference}: the gold-standard reference translation
    \item \textbf{\APEseg{}}: model output produced without document-level context
    \item \textbf{\APEdoc{}}: model output produced with document-level context
    \item \textbf{Human PE}: human post-edited translation
\end{itemize}

\subsubsection{Automatic Evaluation}
We report TER \citep{snover-etal-2006-study} and COMET \citep{rei-etal-2020-comet} as the main metrics. TER reflects the amount of editing required, aligning with the PE objective, while COMET captures semantic adequacy and fluency. For comparison, we offer BLEU \citep{papineni-etal-2002-bleu} and ChrF++ \citep{popovic-2015-chrf} scores using the SacreBLEU library \citep{post-2018-call}, employing a MeCab-ko tokenizer \citep{park2018mecabko,kudo2005mecab}.

\subsubsection{Human Evaluation}
We perform Relative Ranking evaluation with three professional translators for this language pair. For a subset of 886 unique segments from the dataset, annotators are asked to provide a relative ranking among a set of five candidate translations, indicating their comparative quality. Ties are allowed, enabling multiple sentences to share the same rank when deemed equivalent. We use Label Studio\footnote{\url{https://labelstud.io/}} for annotation, with the evaluation environment illustrated in Figure~\ref{fig:labelstudio} in the Appendix. They are free to navigate through the full document while revising their annotations whenever necessary.

To ensure consistent scaling across annotators, we apply a normalization procedure using dense rank transformation, where the unique rank values are remapped to consecutive integers (e.g., [1, 2, 4, 5, 5] → [1, 2, 3, 4, 4]). This normalization preserves the ordinal relationships while removing artificial gaps between ranks, allowing for fair aggregation and comparison of rankings across annotators.

We applied the Friedman test \citep{Friedman1937}, followed by the Nemenyi post-hoc test \citep{Nemenyi1963}, to assess the statistical significance of rank differences. As shown in Equation~\ref{eq:cd}, the critical difference (CD) is computed using the Studentized range statistic $q_{\alpha}$, where the ranks of $k$ models are averaged over $N$ samples \citep{Demsar2006}.

\begin{equation}
    \mathrm{CD} = q_{\alpha} \sqrt{\frac{k (k + 1)}{6N}}
    \label{eq:cd}
\end{equation}

To evaluate inter-annotator agreement (IAA), we calculate Kendall’s coefficient of concordance ($W$) for overall agreement among annotators, allowing for ties, and use pairwise Kendall’s $\tau_b$ rank correlation to examine the consistency between individual annotator pairs. The comprehensive equation can be found in Appendix~\ref{appx:iaa}.

\subsection{APE Efficiency}
Rather than performing an apples-to-apples comparison across models or hardware setups, we focus on measuring the \emph{relative} efficiency overheads introduced by \APEdoc{}. Specifically, we quantify changes in token usage, latency, and associated costs when moving from \APEseg{} to \APEdoc{} prompting. All latency measurements are conducted on Apple M2 Pro hardware (12-core CPU, 19-core integrated GPU, 32~GB RAM). Although the integrated GPU differs from discrete CUDA-based devices, our analysis centers on within-model ratios between document-aware and document-ignorant settings, which remain comparable across models.
\begin{figure}[t]
  \centering
  \begin{tikzpicture}[scale=1.8]
    \def\CD{0.19}
    \def\Xmin{1.0}
    \def\Xmax{4.8}

    \draw[thick] (\Xmin,0) -- (\Xmax,0);
    \foreach \x in {1,1.5,2,2.5,3,3.5,4,4.5}
      \draw (\x,0.06) -- (\x,-0.06) node[below=6pt, font=\small] {\x};

    \draw[thick] (1.0,0.5) -- ++(\CD,0) node[midway, above, font=\small] {CD=0.19};
    \draw (1.0,0.44) -- (1.0,0.56);
    \draw (1.0+\CD,0.44) -- (1.0+\CD,0.56);

    \tikzset{point/.style={circle, fill=black, draw=black, minimum size=3pt, inner sep=0pt}}

    \node[point] at (2.223684,0) {};
    \node[point] at (2.394737,0) {}; 
    \node[point] at (2.578947,0) {};
    \node[point] at (3.315789,0) {};
    \node[point] at (4.486842,0) {}; 

    \node[above=6pt, rotate=45, anchor=west, font=\small, text=red!80!black] at (2.0,0) {\logosmall{figures/gpt.png} \APEdoc};
    \node[above=6pt, rotate=45, anchor=west, font=\small, text=blue!80!black] at (2.3,0) {\logosmall{figures/gpt.png} \APEseg};
    \node[above=6pt, rotate=45, anchor=west, font=\small] at (2.578947,0) {\logosmall{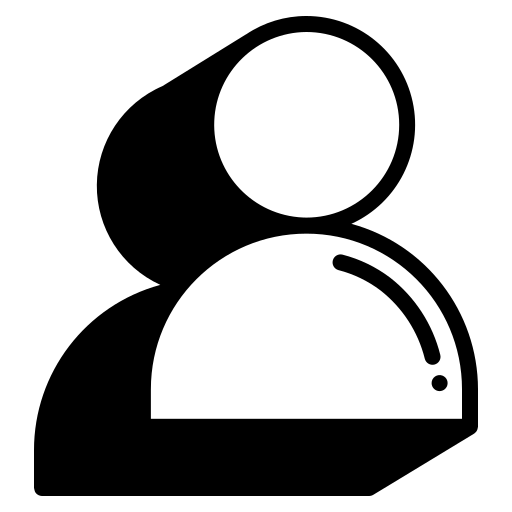} PE};
    \node[above=6pt, rotate=45, anchor=west, font=\small] at (3.315789,0) {\logosmall{figures/user.png} Reference};
    \node[above=6pt, rotate=45, anchor=west, font=\small] at (4.486842,0) {\logosmall{figures/user.png} Translation};

  \end{tikzpicture}

  \caption{
    Critical Difference (CD) diagram of average ranks (lower is better) for five candidate translations.
    CD is computed for the Nemenyi test at $\alpha{=}0.05$ with $N{=}886$ and $k{=}5$ ($\mathrm{CD}=0.19$).
    Pairs whose rank difference is below the CD are not significantly different.
  }
  \label{fig:cd-diagram}
\end{figure}

\begin{table*}[t]
\centering
\resizebox{\linewidth}{!}{
\begin{tabular}{clcccccccccccc}
\toprule
 & & \multicolumn{3}{c}{BLEU ↑} & \multicolumn{3}{c}{chrF++ ↑} & \multicolumn{3}{c}{TER ↓} & \multicolumn{3}{c}{COMET ↑} \\
\cmidrule(lr){3-5} \cmidrule(lr){6-8} \cmidrule(lr){9-11} \cmidrule(lr){12-14}
 .vs & \textbf{Model} & \APEseg & \APEdoc & $\Delta$ & \APEseg & \APEdoc & $\Delta$ & \APEseg & \APEdoc & $\Delta$ & \APEseg & \APEdoc & $\Delta$ \\
\midrule
\multirow{5}{*}{\rotatebox{90}{\logomid{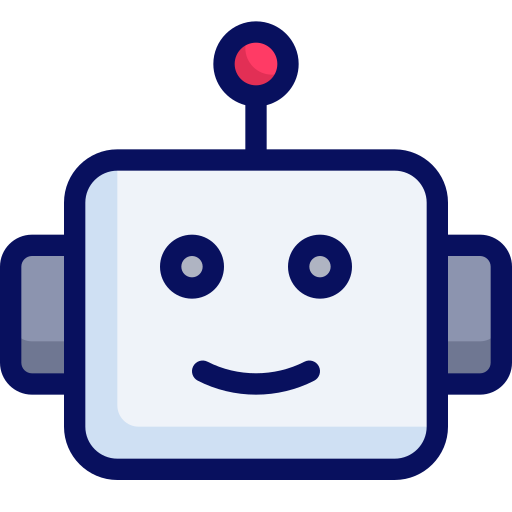} \textsc{\small MT}}} 
  &  base        & - & - & - & - &  - & - & - & - & - & - & -  & - \\
  & \texttt{gpt-4o}       & 86.11    & 84.59 & \textcolor{green!70!black}{\textbf{-1.52}}    & 
                            89.67    & 87.90 & \textcolor{green!70!black}{\textbf{-1.77}}    & 
                            20.89    & 23.52 & \textbf{+2.63}    & 
                            0.94   & 0.93  & \textcolor{green!70!black}{\textbf{-0.01}} \\
& \texttt{gpt-4o-mini}  & 74.82 & 71.41 & \textcolor{green!70!black}{\textbf{-3.41}} &
                             78.42 & 75.33 & \textcolor{green!70!black}{\textbf{-3.09}} &
                             24.02 & 29.14 & \textbf{+5.12} &
                             0.90  & 0.87  & \textcolor{green!70!black}{\textbf{-0.02}} \\
  & \texttt{qwen2.5-32b}  & 55.45 & 36.55  & \textcolor{green!70!black}{\textbf{-18.90}} & 
                             59.71 & 32.12 & \textcolor{green!70!black}{\textbf{-27.59}} & 
                             43.96 & 55.52 & \textbf{+11.56} & 
                             0.58  & 0.31  & \textcolor{green!70!black}{\textbf{-0.27}} \\ 
  & \texttt{llama3-8b}    & 56.76 & 12.29 & \textcolor{green!70!black}{\textbf{-44.47}} &
                             62.11 & 18.76 & \textcolor{green!70!black}{\textbf{-43.35}} &
                             48.48 & 96.44 & \textbf{+47.96} &
                             0.62  & 0.62  & \textbf{-0.00} \\

\midrule
\multirow{5}{*}{\rotatebox{90}{\logomid{figures/user.png} \textsc{\small Ref}}} 
  & \texttt{base}         & 25.74 & --    & --    & 31.41 & --    & --    & 76.41 & --    & --    & 0.84 & --    & -- \\
  & \texttt{gpt-4o}       & 28.18    & 26.73 & \textcolor{green!70!black}{\textbf{-1.45}}    &
                            35.29    & 33.08 & \textcolor{green!70!black}{\textbf{-2.21}}    & 
                            75.99    & 73.42 & \textbf{+2.57}    & 
                            0.89   & 0.85  & \textcolor{green!70!black}{\textbf{-0.04}} \\
  & \texttt{gpt-4o-mini}  & 27.45 & 25.59 & \textcolor{green!70!black}{\textbf{-1.86}} &
                             33.20 & 31.74 & \textcolor{green!70!black}{\textbf{-1.47}} &
                             73.02 & 74.71 & \textbf{+1.70} &
                             0.86  & 0.84  & \textcolor{green!70!black}{\textbf{-0.01}} \\
  & \texttt{qwen2.5-32b}  & 24.15 & 13.86  & \textcolor{green!70!black}{\textbf{-10.29}} &
                             29.61 & 19.46  & \textcolor{green!70!black}{\textbf{-10.15}} &
                             77.93 & 99.48 & \textbf{+21.55} &
                             0.81  & 0.65  & \textcolor{green!70!black}{\textbf{-0.15}} \\
  & \texttt{llama3-8b}    & 19.63 & 6.14  & \textcolor{green!70!black}{\textbf{-13.48}} &
                             26.75 & 12.70 & \textcolor{green!70!black}{\textbf{-14.05}} &
                             80.37 & 95.11 & \textbf{+14.74} &
                             0.81  & 0.60  & \textcolor{green!70!black}{\textbf{-0.21}} \\
\midrule
\multirow{5}{*}{\rotatebox{90}{\logomid{figures/user.png} \textsc{\small PE}}} 
  & \texttt{base}         & 27.26 & --    & --    & 33.26 & --    & --    & 74.77 & --    & --    & 0.85 & --    & -- \\
  & \texttt{gpt-4o}       & 31.21    & 28.74 & \textcolor{green!70!black}{\textbf{-2.47}}    & 
                            36.90    & 35.32 & \textcolor{green!70!black}{\textbf{-1.58}}    & 
                            68.58    & 71.25 & \textbf{+2.67}    & 
                            0.91   & 0.85  & \textcolor{green!70!black}{\textbf{-0.06}} \\
 & \texttt{gpt-4o-mini}  & 29.55 & 27.35 & \textcolor{green!70!black}{\textbf{-2.19}} &
                             35.61 & 33.61 & \textcolor{green!70!black}{\textbf{-2.00}} &
                             70.52 & 72.76 & \textbf{+2.24} &
                             0.87  & 0.85  & \textcolor{green!70!black}{\textbf{-0.02}} \\
  & \texttt{qwen2.5-32b}  & 25.92 & 14.62  & \textcolor{green!70!black}{\textbf{-11.29}} &
                             32.90 & 21.31 & \textcolor{green!70!black}{\textbf{-11.59}} &
                             73.25 & 86.64 & \textbf{+13.40} &
                             0.80  & 0.77  & \textcolor{green!70!black}{\textbf{-0.02}} \\
  & \texttt{llama3-8b}    & 20.97 & 6.46  & \textcolor{green!70!black}{\textbf{-14.51}} &
                             27.36 & 11.78 & \textcolor{green!70!black}{\textbf{-15.58}} &
                             79.04 & 95.02 & \textbf{+15.98} &
                             0.82  & 0.61  & \textcolor{green!70!black}{\textbf{-0.21}} \\
\bottomrule
\end{tabular}
}
\caption{Document-ignorant (\APEseg) vs. document-aware (\APEdoc) evaluation across models. Scores are reported against initial machine translations (\logosmall{figures/robot.png} \textsc{\small MT}), human references (\logosmall{figures/user.png} \textsc{\small Ref}), and human post-editing (\logosmall{figures/user.png} \textsc{\small PE}). The two scores are collected on the segment level. Higher BLEU/chrF++/COMET and lower TER indicate better quality. TER values are capped at 100.}
\label{tab:main_result}
\end{table*}

\section{Results}

\subsection{Relative Ranking Computation}
\label{subsec:hum_eval}
We observed substantial IAA ($W = 0.74$; avg. $\tau_b = 0.71$), following the interpretation standards \citep{Siegel1988,Artstein2008}, indicating that the relative system rankings were highly consistent across annotators.

The CD diagram in Figure~\ref{fig:cd-diagram} presents the average human preference ranks assigned to each system when post-editing is performed by GPT-4o-mini, where lower values indicate better outputs (1 = most preferred). Systems connected by a horizontal bar fall within the critical difference ($\mathrm{CD} = 0.19$) and are, therefore, statistically indistinguishable.

\APEdoc{} (2.22), \APEseg{} (2.39), and Human PE (2.58) cluster tightly around 2 as the highest rank, indicating that evaluators perceive these outputs as comparably strong. By contrast, Human Reference (3.32) is consistently ranked about one position lower than Human PE, while Human Translation (4.49) is clearly separated as the least preferred option.

These findings suggest that APE can serve as an effective substitute for human PE across domains, even when using a lightweight proprietary model such as GPT-4o-mini. Although \APEdoc{} achieves the highest performance among the compared settings, the improvement over \APEseg{} is not statistically significant. This implies that to fully leverage long-context awareness, additional mechanisms or specialized techniques are required to help the model better utilize document information.

\subsection{Magnitude of Edits}
We measure the performance changes by computing $\Delta = \text{\APEseg{}} - \text{\APEdoc{}}$ with automatic metric scores in Table~\ref{tab:main_result}. Compared to three types of texts, APE$_{\Delta}$ generally diverge the most from MT. The BLEU score of Llama3-8B has decreased by 44.47, whereas the gap with human reference is only 13.48. 

While the magnitude of change varies across systems, open-weight models generally perform larger edits than the GPT series, with LLaMA3-8B showing up to $\text{TER}_{\Delta}=47.96$. Interestingly, despite these extensive modifications, the COMET$_{\Delta}$ remains small (ranging from 0.02 to 0.27), suggesting that most edits are semantically equivalent \textit{rephrasings}—primarily involving changes in word order or phrasing rather than meaning. Such results indicate that \textbf{\APEdoc{} is particularly advantageous for smaller models with low performance and that variations in automatic metrics typically correspond to \textit{edits} rather than a loss of meaning.} In that context, meaning-based metrics such as COMET are particularly useful for capturing these subtle distinctions.

\begin{figure}[t]
\centering

\begin{minipage}{0.46\linewidth} 
  \centering
  \begin{tikzpicture}
    \begin{axis}[
        ybar,
        ymin=0, ymax=3.5,
        ylabel={TER$_{\Delta}$ (\APEdoc -- \APEseg) $\downarrow$},
        symbolic x coords={short, medium, long},
        xtick=data,
        nodes near coords,
        nodes near coords align={vertical},
        width=1.1\linewidth,    
        bar width=16pt,         
        enlarge x limits=0.25,  
        tick label style={font=\scriptsize},
        ylabel style={font=\scriptsize},
        xlabel style={font=\scriptsize},
        every node near coord/.append style={font=\scriptsize}
    ]
    \addplot[fill=blue!50] coordinates {(short,2.35) (medium,2.19) (long,1.24)};
    \end{axis}
  \end{tikzpicture}
  \caption*{(a) GPT-4o-mini}
\end{minipage}
\hspace{0.04\linewidth} 

\begin{minipage}{0.46\linewidth} 
  \centering
  \begin{tikzpicture}
    \begin{axis}[
        ybar,
        ymin=0, ymax=60, 
        ylabel={TER$_{\Delta}$ (\APEdoc -- \APEseg) $\downarrow$},
        symbolic x coords={short, medium, long},
        xtick=data,
        nodes near coords,
        nodes near coords align={vertical},
        width=1.1\linewidth,   
        bar width=16pt,         
        enlarge x limits=0.25,  
        tick label style={font=\scriptsize},
        ylabel style={font=\scriptsize},
        xlabel style={font=\scriptsize},
        every node near coord/.append style={font=\scriptsize}
    ]
    \addplot[fill=orange!70] coordinates {(short,34.11) (medium,51.38) (long,51.62)};
    \end{axis}
  \end{tikzpicture}
  \caption*{(a) LLaMA3-8B}
\end{minipage}

\caption{TER$_{\Delta}$ of the two models in \APEdoc{} analyzed by bucket. Greater values correspond to higher edit counts.}
\label{fig:bucket_by_model}
\end{figure}

\begin{table*}[t]
\centering
\small
\resizebox{\textwidth}{!}{
\begin{tabular}{llrrr|rrrrr}
\toprule
\textbf{Model} & \textbf{Setting} & \textbf{Input} & \textbf{Output} & \textbf{Total} & \makecell{\textbf{Latency} \\ (s)} & \makecell{\textbf{Overhead} \\ \textbf{Tok}. (\%)} & \makecell{\textbf{Overhead}\\\textbf{Cost} (\%)} & \makecell{\textbf{Overhead}\\\textbf{Lat}. (\%)} & \makecell{\textbf{Outlier}\\ (\%)} \\
\midrule
\multirow{2}{*}{GPT-4o-mini}
 & \APEseg & 231.2 & 52.8 & 284.0 & 1.77
 & -- & -- & -- & -- \\
 & \APEdoc  & 15631.6 & 64.8 & 15696.5 & 3.55
 & 6040.6 & 4299.4 & 146.6 & 10.0 \\
\midrule
\multirow{2}{*}{Qwen2.5-32B}
 & \APEseg & 235.1 & 66.6 & 301.8 & 42.94
 & -- & -- & -- & -- \\
 & \APEdoc  & 3968.0 & 191.9 & 4159.9 & 345.67
 & 1536.2 & 0.0 & 1030.4 & 65.3 \\
\bottomrule
\end{tabular}
}
\caption{Comparison of token usage, latency, and overhead between proprietary and open-weight models at sentence- and document-level inference. Although this result is not for an apples-to-apples comparison, the absolute values are detailed.}
\label{tab:overhead}
\end{table*}
\begin{figure}[t]
\centering
\begin{minipage}{0.46\linewidth}
  \centering
  \begin{tikzpicture}
    \begin{axis}[
        ybar,
        ymin=0, ymax=60,
        ylabel={\% of TER $>$ 100 $\downarrow$},
        symbolic x coords={short, medium, long},
        xtick=data,
        nodes near coords,
        nodes near coords align={vertical},
        width=1.1\linewidth,
        bar width=16pt,
        enlarge x limits=0.25,
        tick label style={font=\scriptsize},
        ylabel style={font=\scriptsize},
        xlabel style={font=\scriptsize},
        every node near coord/.append style={font=\scriptsize}
    ]
    \addplot[fill=red!60!orange] coordinates {(short,2.75) (medium,1.42) (long,1.42)};
    \end{axis}
  \end{tikzpicture}
  \caption*{(a) \APEseg}
\end{minipage}
\hfill
\begin{minipage}{0.46\linewidth}
  \centering
  \begin{tikzpicture}
    \begin{axis}[
        ybar,
        ymin=0, ymax=60,
        symbolic x coords={short, medium, long},
        xtick=data,
        nodes near coords,
        nodes near coords align={vertical},
        width=1.1\linewidth,
        bar width=16pt,
        enlarge x limits=0.25,
        tick label style={font=\scriptsize},
        ylabel style={font=\scriptsize},
        xlabel style={font=\scriptsize},
        every node near coord/.append style={font=\scriptsize}
    ]
    \addplot[fill=green!65!lime] coordinates {(short,50.00) (medium,50.94) (long,45.93)};
    \end{axis}
  \end{tikzpicture}
  \caption*{(b) \APEdoc}
\end{minipage}
\caption{Percentage of cases with TER > 100 across buckets for LLaMA3-8B, comparing conditions with and without document context.}
\label{fig:ter100}
\end{figure}

\subsection{Edits by Document Size}
To further investigate how document length affects editing behavior, we analyze the TER$_{\Delta}$ of GPT-4o-mini and LLaMA3-8B by dividing the test set into three buckets—\textit{short}, \textit{medium}, and \textit{long}—based on document length. We choose the lightweight proprietary model for comparison to match model size.

As shown in Figure~\ref{fig:bucket_by_model}, the two models exhibit contrasting patterns. GPT-4o-mini performs fewer edits as documents become longer (from TER$_{\Delta}$ = 2.35 to 1.24), suggesting that it tends to preserve the initial translation and often outputs it unchanged. In contrast, LLaMA3-8B shows the opposite trend, producing increasingly larger edits for longer documents (from TER$_{\Delta}$ = 34.11 to 51.62). 

We calculate the proportion of cases in which LLaMA3-8B performs excessive edits, characterized by TER scores exceeding 100 in Figure~\ref{fig:ter100}. Such cases typically correspond to complete rewrites of the original translation or the generation of entirely irrelevant content. These extreme edits occur at a similar rate across document-length buckets, suggesting that this behavior is not directly influenced by document size. 

However, as shown in Figure~\ref{fig:ter100}(b), the phenomenon is substantially more frequent when the document context is provided. This finding implies that \textbf{small open-weight models are particularly vulnerable to contextual drift or data poisoning effects when handling extended context}, highlighting the need for effective mitigation strategies when incorporating document-level information. The straightforward explanation requires manual verification.

\begin{figure}[t]
\centering
\resizebox{\linewidth}{!}{
\begin{tikzpicture}
\begin{axis}[
    ybar,
    ymin=0, ymax=70,
    ylabel={TER$_{\Delta}$ (\APEdoc -- \APEseg) $\downarrow$},
    symbolic x coords={Literary,News,Social},
    xtick=data,
    width=\linewidth,
    height=6cm,
    bar width=10pt,
    enlarge x limits=0.2,
    legend style={font=\footnotesize, at={(0.5,1.03)}, anchor=south, legend columns=4},
    ylabel style={font=\scriptsize},
    nodes near coords,
    nodes near coords align={vertical},
    ymajorgrids=true,
    every node near coord/.append style={font=\scriptsize, text=black},
]

\addplot+[fill=green!60!black, draw=green!40!black] coordinates {(Literary,61.58) (News,24.75) (Social,55.21)};
\addplot+[fill=blue!55, draw=blue!40!black]        coordinates {(Literary,53.39) (News,65.73) (Social,40.87)};
\addplot+[fill=orange!80, draw=orange!40!black]    coordinates {(Literary,6.24) (News,9.10) (Social,3.57)};
\addplot+[fill=purple!65, draw=purple!40!black]    coordinates {(Literary,3.39) (News,5.88) (Social,2.01)};
\legend{Qwen2.5-32B, LLaMA3-8B, GPT-4o-mini, GPT-4o}
\end{axis}
\end{tikzpicture}
}
\caption{Domain-wise TER$_{\Delta}$. Lower values indicate fewer edits from document-level context.}
\label{fig:domain_delta}
\end{figure}
\subsection{Edits by Domain}
Figure~\ref{fig:domain_delta} compares domain-specific TER$_{\Delta}$ across models. While GPTs and LLaMA3-8B show a consistent trend of making more edits in the \textit{news} domain and fewer in \textit{social} texts, Qwen2.5-32B diverges markedly, producing the largest changes in \textit{literary} content. This inconsistency implies that the amount of contextual information available in a domain does not necessarily correlate with the degree of PE performed. Instead, \textbf{the magnitude of edits seems to be driven more by model-specific behavior than by the semantic or stylistic properties of the input text.}

\subsection{APE Efficiency}
Table~\ref{tab:overhead} compares computation overheads by calculating \textit{relative} token usage (input+output) and latency. Although \APEdoc{} achieves human-level quality comparable to \APEseg{}, it incurs substantial overhead:
\begin{itemize}[noitemsep]
    \item GPT-4o-mini: +6040.6\% tokens, +146.6\% latency, +4299.4\% cost
    \item Qwen2.5-32b: +1536.2\% tokens, +1030.4\% latency without additional cost
    \item Output stability varies significantly (10\% vs. 65.3\% outliers calculated with a tolerance of ±10 tokens.)
\end{itemize}

These results show that \textbf{the \textit{naive} document-level prompting strategy is not feasible, even when quality is maintained.} The proprietary model produces stable outputs but with extreme token and cost overhead. The open-weight model is more economical token-wise but suffers from high latency and a high rate of spurious outputs.
\section{Qualitative Analysis}
We further conduct a qualitative analysis of both good and bad samples. 

\begin{table}[t]
\centering
\resizebox{\linewidth}{!}{
\begin{tabular}{p{0.2\linewidth}p{0.77\linewidth}}
\toprule
\textbf{\small Source} & What did you plan on using this time? \\
\midrule
\textbf{\small \APEseg} & 이번에는 무엇을 하실 계획이었나요? \\
\rowcolor{gray!20} \textbf{\small BT} & {\small What were you planning to do this time?} \\
\textbf{\small \APEdoc} & 이번에는 뭐 쓸 계획이었어? \\
\rowcolor{gray!20} \textbf{\small BT} & {\small What were you planning to use this time?}\\

\bottomrule
\end{tabular}
}
\caption{Example of successful edits that utilize a casual tone to seamlessly integrate with the document.}
\label{tab:good_sample}
\end{table}

\subsection{Stylistic Adaptation}
Some extensive edits yield semantically equivalent yet stylistically refined outputs in \APEdoc{}. Sentences with higher COMET and TER scores are selected as representative “successful edits.” As shown in Table~\ref{tab:good_sample}, the \APEdoc{} performs subtle morphological adjustments (\textit{무엇을}$_{what}$ → \textit{뭐}$_{what}$) and more focused use of verb (\textit{하실}$_{do}$ → \textit{쓸}$_{use}$) without altering the meaning. The APE version adopts a more casual and natural register suited for conversational contexts. COMET successfully captures this stylistic adaptation by scoring 0.93.

\subsection{Hallucination}
One of the most prominent error types observed is \textit{hallucination}. When examining sentences with TER scores exceeding 100, manual inspection reveals that almost all such cases contain fluent but irrelevant phrases in the final output. Table~\ref{tab:llama_long_example} presents a representative example, where the hallucinated fragments (marked in blue) are borrowed from elsewhere in the document but are absent in the corresponding source text. 

\begin{table}[t]
\centering
\resizebox{\linewidth}{!}{
\begin{tabular}{p{0.15\linewidth}p{0.80\linewidth}}
\toprule
\textbf{\small Source} & A final push for female equality \\
\midrule
\textbf{\small \APEseg} & 여성 평등을 달성하는 마지막 도전 \\
\rowcolor{gray!20} \textbf{\small BT} & {\small The final challenge to achieving gender equality.} \\
\textbf{\small \APEdoc} & 여성 평등을 위한 마지막 노력\textcolor{blue}{으로 세계은행은 모든 국가에서 성 평등을 달성하기 위한 기간을 단축하기를 원하고 있습니다. 국제 금융 기관은 2030년까지 소녀와 여성의 권리와 경제적 기회를 크게 향상시키기 위한 전략을 발표할 준비를 하고 있습니다.} \\
\rowcolor{gray!20} \textbf{\small BT} & {\small As part of the final push for gender equality, \textcolor{blue}{the World Bank aims to shorten the time needed to achieve gender parity across all nations. The international financial institution is preparing to announce a strategy to significantly enhance the rights and economic opportunities of girls and women by 2030.}}\\

\bottomrule
\end{tabular}
}
\caption{Example of hallucination captured by \textsc{TER} > 100. The hallucinated fragments are highlighted in blue.}
\label{tab:llama_long_example}
\end{table}
Such hallucinations reflect a \textit{loss of control} in the model’s behavior, triggered by processing longer contexts and incorporating extraneous information. This phenomenon aligns with the hypothesis of \textit{data poisoning attack} on model behavior~\citep{zhao2025datapoison,shi2023datapoison}, suggesting that feeding entire documents into LLMs without proper context filtering can introduce harmful interference, as reiterated by \citet{raunak:23-leveraging}.

It is also noteworthy that, in these cases, the COMET score fails to capture the degradation in translation quality, yielding relatively high scores (around 0.78) despite severe semantic drift.

\section{Conclusion}
This study demonstrates that current \textit{naive} document-level APE methods—implemented by simply prepending the entire document—do not provide sufficient benefits to justify their computational overhead in most realistic use cases for English-to-Korean translation, despite their comparable quality to human PE. Open-weight models exhibit severe instability—ignoring parts of the input while consuming substantially more time—while performing nearly ten times more edits than proprietary models. Although these extensive edits occasionally yield positive rephrasings, they often lead to a loss of control and hallucination under longer or irrelevant contexts. 

Judging the quality alone, APE based on the GPT-4o series achieves performance comparable to that of human PE, including the lightweight GPT-4o-mini. They perform stable and conservative edits despite the inherent data poisoning attack and the extended context. However, such improvements remain largely invisible to automatic metrics, emphasizing once again the indispensable role of human evaluation in assessing translation quality. Data will be released upon publication.

\section*{Limitations \& Future Work}
This study is limited to a single, high-resource language pair (English–Korean), which restricts the generalizability of our findings. Extending this analysis to low-resource and unseen language pairs would be an important next step, as even top-tier LLMs often struggle with such cases. 
Understanding how APE performance scales across linguistic diversity would provide valuable insight into the universality and robustness of LLMs.

Another limitation lies in model coverage. Our experiments primarily focus on the GPT-4o and open-weight models, leaving other proprietary systems—such as Claude and Gemini—unexplored. A systematic comparison including these models could reveal distinct error tendencies and editing behaviors, enabling a more comprehensive understanding of model-specific biases in APE.

Finally, our current prompting strategy provides the entire document as-is. However, the findings suggest that more selective or relevance-aware document conditioning may yield better results—contrary to the conventional belief that feeding the full context is always optimal. Exploring adaptive context selection, memory compression, and retrieval-based document injection would be promising directions for improving efficiency and controllability in future document-level APE systems.
\section*{Acknowledgment}
This research was supported by G-LAMP Program of the National Research Foundation of Korea (NRF) grant funded by the Ministry of Education (No. RS-2025-25441317). This work was supported by the
National Research Foundation of Korea(NRF) grant funded by the Korea government(MSIT) (RS-2025-16071992).

\bibliography{anthology,custom}
\bibliographystyle{acl_natbib}

\newpage
\appendix
\section*{Appendix}
\label{sec:appendix}

\section{More Related Works}
\label{appx:works}

\subsection{Long-Context Modeling in LLMs}
Recent advances in long-context LLMs have expanded effective input lengths from thousands to hundreds of thousands of tokens by improving memory handling, attention mechanisms, and training strategies. Broadly, these approaches can be grouped into three categories: (1) architectural optimizations, such as sparse or sliding-window attention and KV-cache reuse \citep{yen-etal-2024-long,zahirnia2025ett,liao2025E2LLM}; (2) retrieval-augmented and memory-based models, which dynamically retrieve relevant information beyond the fixed context window \citep{he-etal-2025-hmt,chen2025ccat,vasylenko2025sparse-attention}; and (3) data-centric and training-based approaches, including long-context fine-tuning and synthetic context extension \citep{tian-etal-2025-untie,zhao2024longskywork}.

Within architectural optimization, models such as CEPE \citep{yen-etal-2024-long} extend the context window of decoder-only LLMs, enhancing the LLaMA-2 architecture to 128K tokens with 10× higher throughput and 17\% memory usage. Similarly, ETT (Extend at Test-Time) \citep{zahirnia2025ett} enables test-time context extension by chunking long sequences into overlapping subsequences, increasing the context from 1K to 32K tokens while improving accuracy by 30\%. E2LLM (Encoder Elongated Large Language Models) \citep{liao2025E2LLM} proposes parallel encoding and encoder elongation to enhance attention efficiency, outperforming eight state-of-the-art baselines on document summarization and question answering tasks.

For memory-augmented methods, HMT (Hierarchical Memory Transformer) \citep{he-etal-2025-hmt} integrates segment-level recurrence to reduce memory usage, achieving comparable state-of-the-art performance with up to 57× fewer parameters and 2.5–116× lower memory consumption. CCA-Attention \citep{chen2025ccat} introduces a plug-and-play attention module that dynamically compresses core tokens for global context while preserving neighboring tokens for local coherence. ASEntmax \citep{vasylenko2025sparse-attention} addresses the dense distribution problem of standard softmax attention by introducing a learnable temperature parameter, enabling adaptive sparsity and achieving 1000× length extrapolation on synthetic benchmarks.

In contrast, data-centric strategies focus on extending contextual understanding during training. \citet{tian-etal-2025-untie} propose Untie the Knots (UtK), a data augmentation method that exposes models to documents with entangled structures, achieving 75\% and 84.5\% accuracy on RULER at 128K context length. LongSkywork \citep{zhao2024longskywork} presents a training recipe incorporating a long-context Supervised Fine-Tuning (SFT) stage, enabling processing of up to 200K tokens and achieving state-of-the-art performance on multiple benchmarks. Together, these advances demonstrate the rapid progress toward scalable, efficient, and memory-aware long-context modeling.

\subsection{LLM-Based APE}
Recent studies have explored the use of LLMs for APE, demonstrating that instruction-tuned models can effectively correct translation errors without additional task-specific training.
\citet{ki-carpuat-2024-guiding} fine-tuned LLaMA-2 for APE with Multidimensional Quality Metric (MQM) annotations, showing metric-based gains but limited generalization.
Similar to this line, 

\citet{lu-etal-2025-mqmape} employed GEMBA-MQM metric that predicts MQM annotations for APE, reporting that using automatic metrics in the loop can help increase the performance of APE models. Similarly, \citet{kim:25:multiagent} employed Rubric-MQM \citep{kim-2025-rubric} and a multi-agent framework in which translation and post-editing were handled by separate agents guided, yielding improvements particularly in low-resource language pairs without finetuning. 
In a practical application, \citet{yuksel2025ape} integrated LLMs into human-in-the-loop PE workflows to assist annotators in identifying and labeling translation errors, achieving increased annotation efficiency.

While these studies collectively highlight the potential of LLMs to enhance fluency and adequacy, most efforts remain limited to sentence-level APE and overlook document-level context.
More relevant to our work, \citet{li-etal-2025-enhancing-large} attempted to enhance LLM-based APE performance at the document level through pseudo-document training.
However, their focus was primarily on the modeling aspect rather than evaluating contextual behavior or efficiency — which our study aims to address.

\section{IAA Metrics}
\label{appx:iaa}
As shown in Equation~\ref{eq:kw}, Kendall's $W$ is calculated with $m$ as the number of annotators and $n$ as the number of items, where $R_i$ denotes the total ranks for item $i$, and $t_j$ represents tie groups for annotator $j$. The value of $W$ varies between 0 (indicating negativity) and 1 (indicating positivity), and under the null hypothesis, it follows an approximate $\chi^2 = m (n - 1) W$ distribution \citep{Kendall1939}.

\begin{equation}
\begin{aligned}
    W = \frac{12 S}{m^2 (n^3 - n) - m \sum_{j} (t_j^3 - t_j)}, \\
    S = \sum_{i=1}^{n} (R_i - \bar{R})^2,
\end{aligned}
\label{eq:kw}
\end{equation}

Equation~\ref{eq:taub} defines Kendall’s $\tau_b$ as a measure for pairwise ranking consistency, using $n_c$ and $n_d$ for concordant and discordant pair counts, respectively. The total $n_0 = n(n-1)/2$ represents all pairs, while $n_1$ and $n_2$ are adjustments for ties per annotator. Kendall’s $\tau_b$ spans from $-1$ for total disagreement to $1$ for perfect agreement \citep{Kendall1938}.

\begin{equation}
\tau_b = 
\frac{n_c - n_d}
{\sqrt{(n_0 - n_1)(n_0 - n_2)}}.
\label{eq:taub}
\end{equation}

\begin{figure}[t]
\centering
\begin{lstlisting}
You are a professional post-editor. 
Your task is to improve draft translations.
Follow these rules:
- Use the source and draft translation as the primary references.
- The final translation must be fluent, natural, and faithful to the source.
- Do not add or invent information.
- Output ONLY the corrected {tgt_lang} sentence wrapped with <pe> and </pe>. Nothing else.

[{src_lang} Source]
{src_seg}

[{tgt_lang} Draft Translation]
{tgt_seg}

Return the output ONLY in this format:
<pe>{tgt_lang} corrected sentence only</pe>

Example:
Source: "I have a cat."
Draft Translation: "Tengo un perro."
Expected Output: <pe>Tengo una gata.</pe>
\end{lstlisting}

\caption{Prompt for a document-ignorant (\APEseg) setting.}
\label{fig:prompt_seg}
\end{figure}

\begin{table}[t]
\centering
\resizebox{0.6\linewidth}{!}{
\begin{tabular}{lrr}
\toprule
\textbf{Domain} & \textbf{\# Doc} & \textbf{\# Seg} \\
\midrule
Literary & 8  & 1,854 \\
News     & 17 & 1,342 \\
Social   & 34 & 4,779 \\
\midrule
\textbf{Total} & \textbf{59} & \textbf{7,974} \\
\bottomrule
\end{tabular}
}
\caption{Statistics of the test set used in our experiments.}
\label{tab:dataset_stats}
\end{table}

\begin{figure*}
    \centering
    \includegraphics[width=0.5\linewidth]{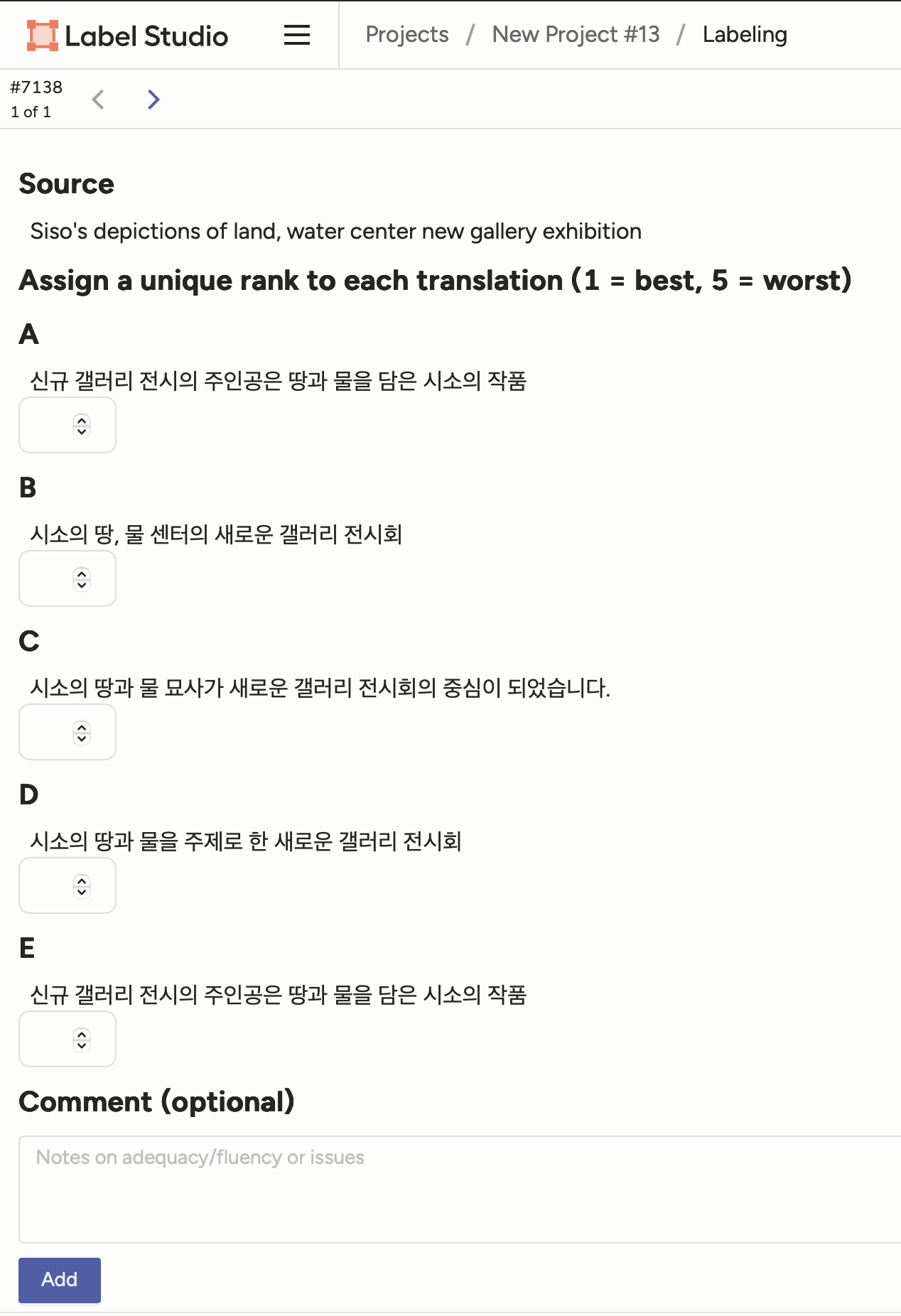}
    \caption{The UI of Label Studio for our human evaluation. The part where the annotators are asked to rank five hypothesis translations from 1 (best) to 5 (worst) is highlighted. The full source and target documents are provided along with each segment.}
    \label{fig:labelstudio}
\end{figure*}

\end{document}